\documentclass[runningheads]{llncs}
\usepackage{graphicx}

\usepackage{tikz}
\usepackage{comment}
\usepackage{amsmath,amssymb} 
\usepackage{color}

\usepackage[accsupp]{axessibility}  

\usepackage{subfigure}
\usepackage{multirow}
\usepackage{hyperref}

\begin{document}
\pagestyle{headings}
\mainmatter
\def\ECCVSubNumber{5044}  

\title{AssistQ: Affordance-centric Question-driven Task Completion for Egocentric Assistant} 

\titlerunning{Affordance-centric Question-driven Task Completion}
\authorrunning{Affordance-centric Question-driven Task Completion} 
%
\author{Benita Wong\thanks{Equal Contribution. $^\dag$ Corresponding Author.} \and
Joya Chen$^*$ \and
You Wu$^*$ \and
Stan Weixian Lei \and
\\
Dongxing Mao \and
Difei Gao \and
Mike Zheng Shou$^\dag$}

\institute{Show Lab, National University of Singapore \\
\email{benitawong@u.nus.edu}
\email{\{joyachen97,mike.zheng.shou\}@gmail.com}
}
\maketitle

\begin{abstract}
A long-standing goal of intelligent assistants such as AR glasses/robots has been to assist users in affordance-centric real-world scenarios, such as \textit{``how can I run the microwave for 1 minute?''}. However, there is still no clear task definition and suitable benchmarks. In this paper, we define a new task called Affordance-centric Question-driven Task Completion, where the AI assistant should learn from instructional videos 
to provide step-by-step help in the user's view. To support the task, we constructed AssistQ, a new dataset comprising 531 question-answer samples from 100 newly filmed instructional videos. We also developed a novel Question-to-Actions (Q2A) model to address the AQTC task and validate it on the AssistQ dataset. The results show that our model significantly outperforms several VQA-related baselines while still having large room for improvement. We expect our task and dataset to advance Egocentric AI Assistant's development. Our project page is available at: \url{https://showlab.github.io/assistq/}.
\end{abstract}

\section{Introduction}\label{introduction}
\begin{figure}[t]
    \centering
    \includegraphics[width=\linewidth]{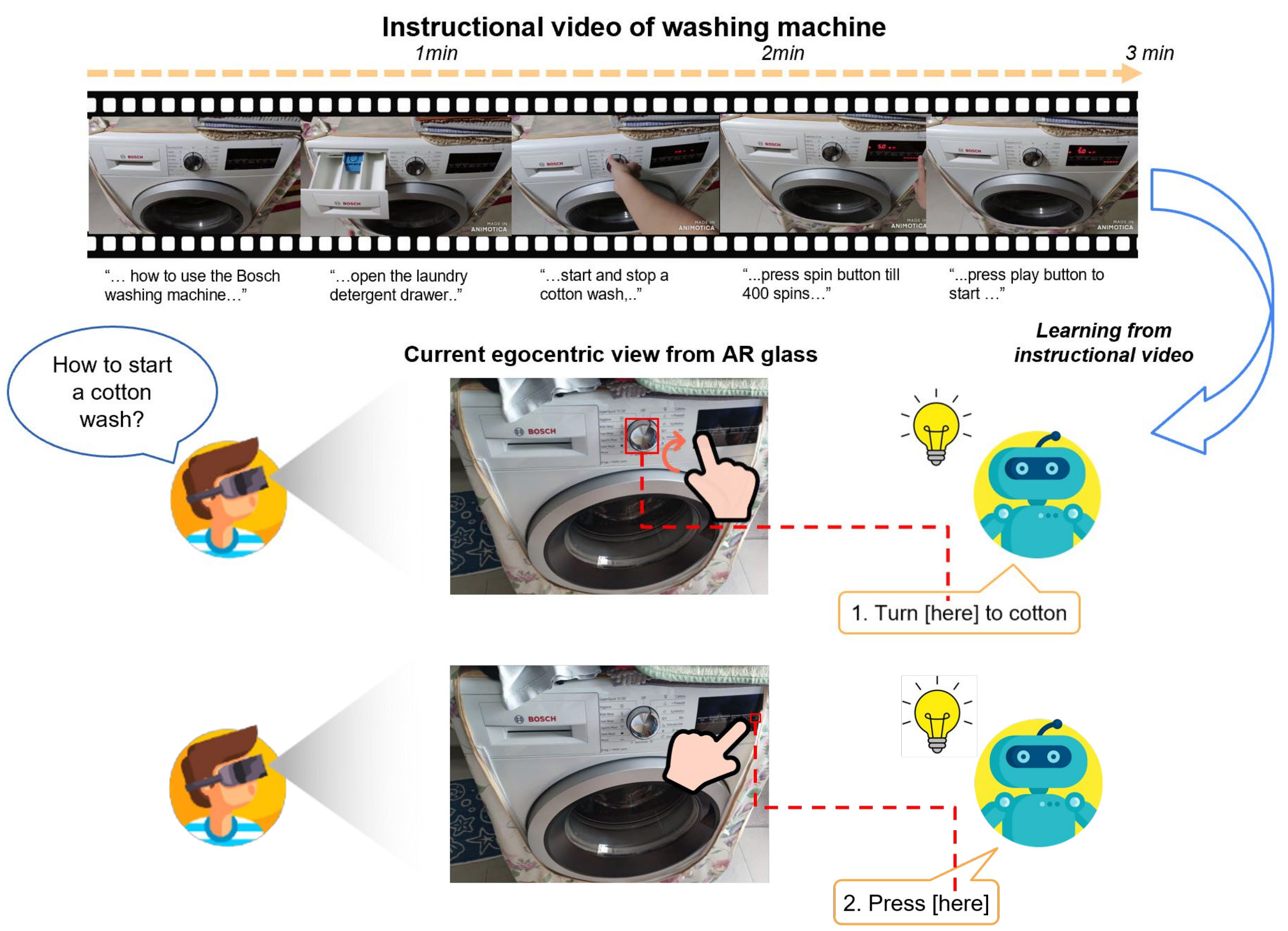}
    \caption{An illustration of an AI assistant for affordance-centric questions. The AI assistant on AR glass can guide the user to complete the intended task.}
    \label{figure1}
\end{figure}

People often require assistance when dealing with new events. Consider the example in Figure~\ref{figure1}: the user comes across an unfamiliar washing machine and wants to start a cotton wash, but he does not know how to operate it. He may search for the device's instructional video, and experiment with buttons on the machine. These actions are time-consuming and may not address the user's question effectively. This example highlights the need for an intelligent assistant to help us with affordance-centric queries. The intelligent assistant should: (1) understand the user's query and view, (2) learn from instructional video/manual, (3) guide the user to achieve his goal.

However, few works support the development of such an assistant. First, AI assistants should deal with affordance-centric queries, while current visual question answering (VQA) benchmarks~\cite{vqa,envqa,vqav2,tgifqa,tvqa,movieqa} focus on fact-based questions and answers. Second, most ground-truth answers in VQA benchmarks are single-step and textual, which fails to express detailed instructions to the user's complex question. Even though visual dialog~\cite{viddial,visdial} (VDialog) introduces sequence question answering, the answer of each dialog round is still single-step and textual. Finally, AI assistants should solve questions in users' egocentric perspective~\cite{ego4d,egovlp,ego_visualprior,ego_nav,ego_scenepriors}, which is more natural and meaningful for understanding users' situations. However, existing VQA and VDialog studies mainly comprise third-person web videos. Although embodied QA~\cite{eqa,mt_eqa} (EQA) attempts to develop egocentric agents, they are mostly focused on virtual environments, which may be difficult to generalize to real-world scenarios, and they are also incapable of dealing with complex, affordance-centric user queries.

\begin{figure}[t]
    \centering
    \includegraphics[width=\linewidth]{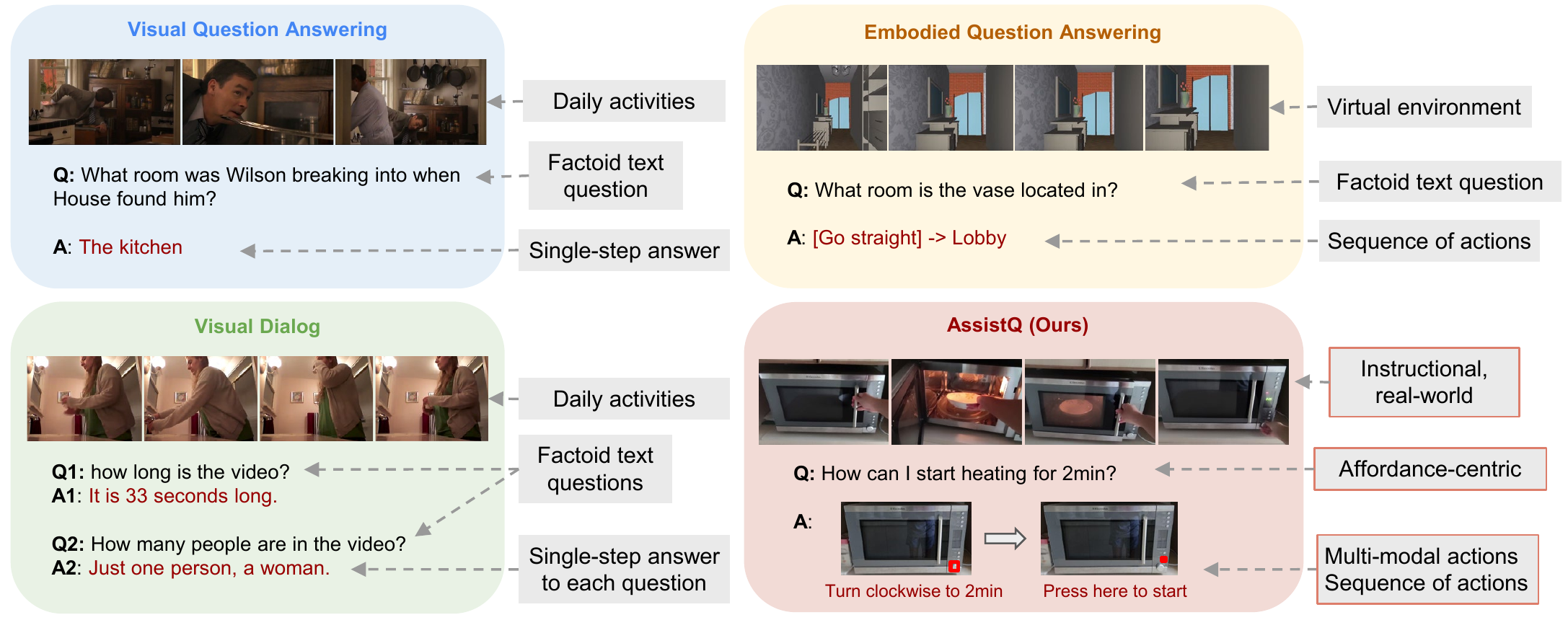}
    \caption{The differences between VQA, VDialog, EQA, and our AQTC task.}
    \label{figure2}
\end{figure}

Hence, we propose a novel \textbf{Affordance-centric Question-driven Task Completion (AQTC)} task. It presents a unique set-up: the user asks a question in his/her view, and then the AI assistant answers in a sequence of actions. Fiugre~\ref{figure2} shows the differences between our task and VQA, VDialog, and EQA. Unlike VQA and VDialog, where the question is fact-based and the answer is a short text, our question is task-oriented and our answer is multi-modal and multi-step, which is more challenging and closer to the intelligent assistant setting. Furthermore, our task makes use of new content -- real-world, egocentric videos -- for question answering, which is more suitable for real-world applications compared to virtual environments in the EQA.

To support the AQTC task, we collected the AssistQ dataset containing 531 multiple-choice QA samples on 100 newly filmed, with an average duration of 115s egocentric videos. Participants recorded instructional videos by themselves (\textit{e.g.,} operating microwaves, washing machine), and provided scripts of their narration. These instructional videos were used to create affordance-centric, multi-step QA samples. We also developed a Question-to-Actions (Q2A) model to address the AQTC task. It uses a context grounding module to infer from multi-modal cues (video, script, user view), and introduces a steps network to generate step-by-step answers. Experimental results show that our Q2A model is especially suitable for the AQTC task. In summary, our contributions are:

$\bullet$ We proposed a new task, namely Affordance-centric Question-driven
Task Completion (AQTC), whereby AI assistants should learn from instructional videos to guide users step-by-step in the users' view.

$\bullet$ We constructed a pilot dataset for the task, comprising 531 question-answer samples derived from 100 newly filmed instructional videos.

$\bullet$ We developed a Question-to-Actions (Q2A) model and performed extensive experiments to show its superiority in the AQTC task.

\section{Related Work}\label{related_work}

\noindent\textbf{Visual Question Answering.} The goal of VQA is to answer questions based on visual cues. VQA v1 and v2~\cite{vqa,vqav2} are standard benchmarks for image QA. TGIF-QA~\cite{tgifqa} is one of the earlier large-scale video QA datasets and was built from short animated videos (GIFs) taken from Tumblr. In this task, the main goal is to recognize actions and their repetitions. Similar video QA datasets~\cite{envqa,tvqa,tvqa+,movieqa} mainly illustrate daily life or story. In contrast, our task focuses on instructional videos more suitable for affordance-centric tasks.

\noindent\textbf{Visual Dialog.} VDialog~\cite{visdial} is a task to generate natural responses based on the given image/video and the dialog context. An example is Audio Visual Scene-Aware Dialog (AVSD)~\cite{viddial}, where the agent is shown a video of indoor activity and answers multiple questions about actions and events in the video. Compared to models developed for single-step VQA tasks, VDialog models should utilize the dialog history. Some models~\cite{viddial,visdial,rlm,simple_viddial} simply splice historical dialogs or model using RNN, while others~\cite{rl_viddial,nmn_viddial} introduce reinforcement learning methods. Like VDialog, historical actions in our AQTC task may affect the current answer step. But their differences are apparent: Visual Dialog is more like a single-step QA problem in each round, whereas AQTC requires a sequence of actions to answer. We believe the latter is more appropriate for affordance-centric questions.

\noindent\textbf{Embodied Question Answering.} If we consider the instructional video in our AQTC task as the environment, then our model should explore the environment to answer the question. This is similar to EQA~\cite{eqa}, where the agent needs to explore the scene and answer the question. Recent works try to solve or extend EQA by improving the logical reasoning~\cite{nmc,mt_eqa}, applying multi-agent cooperation~\cite{tarmac,2body,mt_eqa}, or creating more challenging environments~\cite{iqa}. However, the current setting of EQA focuses on virtual environments, while our proposed AQTC task revolves around real-world images and videos. This allows AQTC to be more readily adopted for practical applications, such as intelligent agents built in portable devices (\textit{e.g.,} mobile phones, AR glasses).

\section{Affordance-centric Question-driven Task Completion}\label{task}

In this section, we propose Affordance-centric Question-driven Task Completion (AQTC), a novel task aimed at driving the evolution of egocentric AI assistants. We begin by outlining the abilities required of an egocentric, user-helping AI assistant. The AQTC work is then formalized in accordance with these abilities. Finally, we talk about how to generalize and apply the task.

\subsection{Required Abilities of Egocentric AI Assistants}\label{section3.1}

\noindent\textbf{Handling multi-modal inputs.} In affordance-centric settings, the user often asks questions in both visual (\textit{e.g.,} his/her current view) and linguistic (\textit{e.g.,} \textit{"how to start the microwave?"}). As a result, the egocentric AI assistant is expected to handle multimodal inputs.

\noindent\textbf{Learning from instructional sources.} It is difficult for an AI assistant to answer questions without learning anything. An acceptable way is for the AI assistant to read the instructional manual first and then teach users. We notice that there are various instructional videos available on the internet, and it would be preferable if AI assistants could utilize these data, \textit{i.e.,} learning from instructional videos and then guide users in their situation.

\noindent\textbf{Grounding between modalities.} The different input modalities are often not grounded with respect to each other. For example, a video guide might point to a button, whereas the corresponding textural guide may say ``Press this''. The egocentric AI assistant should know how to link visual and language concepts.

\noindent\textbf{Multi-step interaction with users.} Interaction with the user is necessary for an egocentric AI assistant. For example, if a user inquires about \textit{"how to start the microwave,"} the assistant should tell him/her the location of the power switch. When a user's inquiry is complex, the interaction between the user and the assistant may be held for several steps.

\subsection{Task Definition of AQTC}\label{section3.2}

Now we design a task for the abilities discussed in Section~\ref{section3.1}. To begin, the task should include multi-modal inputs, such as the user's egocentric view and textural question. Then, there should be a reference instructional video from which AI assistants can learn to address problems. Furthermore, rather than using different modalities independently, the grounding between them should be necessary to handle the task. Finally, the task should reflect multi-step interactions between users and assistants, such as a sequence of actions taught to users by the assistant.

With these considerations, we propose affordance-centric question-driven task completion (AQTC) task. Given an instructional video ($V$), the corresponding video's textural script ($S_V$), the user's view ($U$), and the question ($Q_U$) asked from the view, the model should answer the question by a sequence of answers $(A^1, A^2, ..., A^I)$ in $I$ ($I >= 1$) steps. Note that the question $Q_U$ is proposed under the user view $U$ rather than the instructional video $V$, so the model should ground between $U$ and $V$ (with $S_V$) to answer $Q_U$. Currently, we set candidate answers (${A_j^i}$) to lower the difficulties of answering, where $A_j^i$ denotes the $j$-th potential answer in the $i$-th step.

Comparing with existing VQA-related tasks (\textit{e.g.,} VQA~\cite{vqa,tvqa,movieqa}, VDialog~\cite{viddial,visdial}, EQA~\cite{eqa}), our proposed AQTC task is more suitable in supporting the affordance-centric requirements of AI assistants. We list the reasons in Table~\ref{table0}:

\textbf{(1)} Property: existing VQA-related datasets are mostly factoid, but some applications go beyond just visual facts and focus on affordance. 

\textbf{(2)} Model Input: our work is the first to consider both reference instructional video and the video of user's current situation.

\textbf{(3)} Model Output: to complete the user's task, our model needs to operate across multiple steps, necessitating the model to perform an action described by textual answer at each step.

\begin{table}[t]
\scriptsize
\centering
\begin{tabular}{c|c|c|c}
Tasks &  Property & Model Input & Model Output \\
\hline
Video QA & Factoid & Reference Video  & Single textual answer \\
Video Dialog & Factoid & Reference Video & Multi textual answers\\
Embodied QA & Factoid & User Situation & Multi actions, Single textual answer  \\
\hline
AQTC & Affordance & Reference Video, User Situation & Multi $<$action, textual answer$>$ pairs \\
\end{tabular}
\caption{Comparisons with VQA-related tasks.}\label{table0}
\end{table}

\subsection{Task Application of AQTC}

Our AQTC task actually proposes a typical and general AI process to assisting humans, \textit{i.e.,} model solves a user's query depicted in text and user view image by seeking answers from the instructional video. Therefore, the task can support a wide range of applications:

\textbf{(1)} Life assistant: supporting AR glass to guide users in daily life, like operating home appliances, cooking dishes. 

\textbf{(2)} Skills training: reducing the labor costs, \textit{e.g.,}, replacing human instructors to teach bicycle repair/PC assembly. 

\textbf{(3)} Working partner: improving people's working efficiency, such as recent \href{https://www.autoevolution.com/news/darpa-wants-ai-assistants-in-black-hawks-first-one-called-ocarina-180524.html}{DARPA's Virtual Partners program}. In research, our task is also not limited in a specific scope. It can be a typical task in egocentric AI, which is a new trend in computer vision (\textit{e.g.,} Ego4D dataset).

\section{AssistQ Benchmark}\label{dataset}
\begin{figure}[t]
    \centering
    \includegraphics[width=\linewidth]{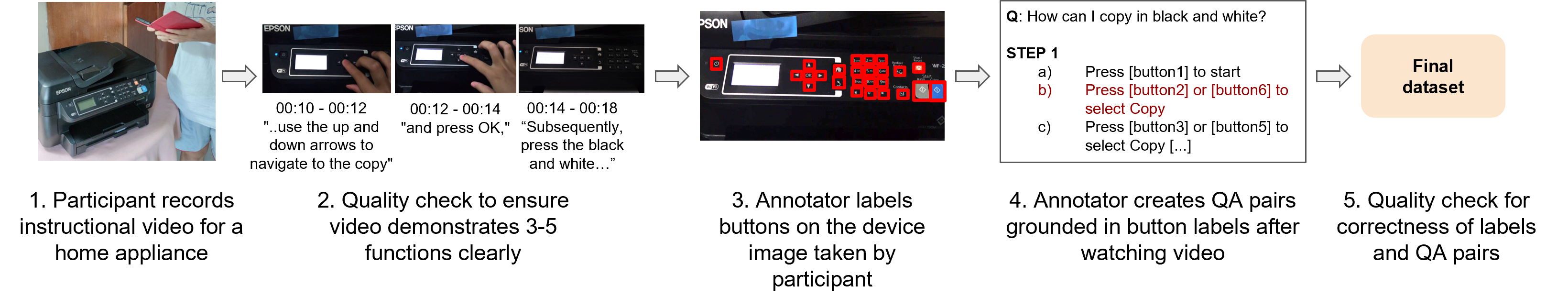}
    \caption{Data Collection Pipeline for our AssistQ benchmark.}
    \label{pipeline}
\end{figure}

We now describe our new benchmark for the AQTC task. The data collection and annotation pipeline is shown in  Figure~\ref{pipeline}. Participants would record videos of themselves operating home appliances, and then annotators would create QA pairs after watching the instructional video. 

\subsection{Video Collection}
\textbf{Videos.} The recruited participants were asked to record instructional videos for home appliances. In each video, the participant demonstrated how to operate a device, such as a washing machine or oven. Participants were asked to propose the functions that they were to record to ensure that the video was sufficiently informative for meaningful question-answering. For example, filming the steps needed to change the temperature of a microwave would result in an information-rich video of interest to AssistQ, whereas filming a video about merely turning the device on would be inadequate. Participants were also encouraged to perform at least 3-5 functions in each video and were given video samples on the recommended placement of their camera to ensure that the videos recorded are of a baseline quality. In addition, participants were required to narrate their actions in the video similar to the audio instructions that accompany product videos which were then submitted.

\subsection{Query Collection}

\begin{figure}[t]
    \centering
    \includegraphics[width=0.95\linewidth]{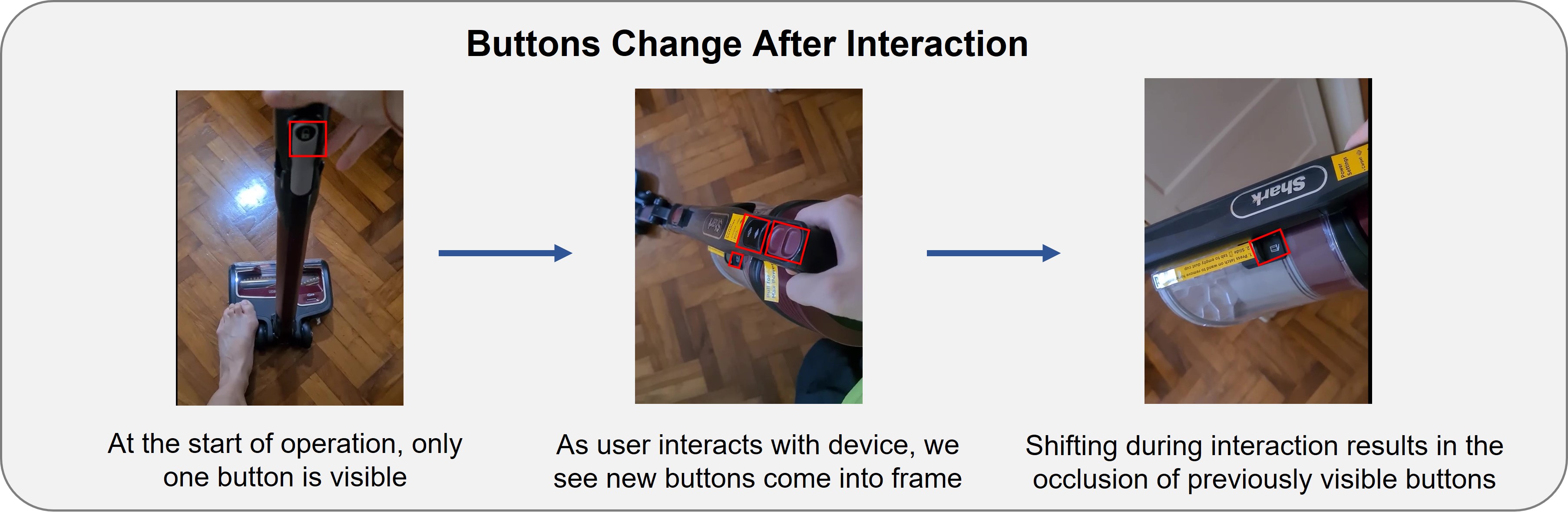}
    \caption{Sample Device (vacuum cleaner) with new buttons in the video frame.}
    \label{Vacuum}
\end{figure}

\textbf{User images and Button Annotation.} For visual query, participants took front-facing pictures of their appliances. In most cases, one image was sufficient to capture all device buttons. But when there is huge change of the device state (Figure~\ref{Vacuum}), participants were asked to provide pictures of the other states to ensure that the agent has sufficient visual information about available choices at each query stage. Annotators were tasked to place bounding boxes over all visible buttons in the image(s), using \href{https://www.makesense.ai/}{Makesense.AI annotation tool}.

\noindent\textbf{QA pairs.} Annotators were asked to create at least 3 questions that were answerable from the video. For greater language diversity in the question set, annotators were encouraged to vary the question stems (\textit{``How do I"},\textit{``How to"},\textit{``What to do to"}), action words (\textit{``cook"},\textit{``bake"}, \textit{``turn''}) and question type. Questions were categories as either specific functionality or pure functionality. Specific functionality questions include specific end-stage values, whereas pure functionality type questions only reference the intended functionality. For example, queries such as \textit{``How to change timer on rice cooker to 3 min"} are specific functionality in nature as they contain a desired end-stage value (3 min), whereas queries such as \textit{``How do I access the timer on the rice cooker"} are pure functionality.

Annotators were asked to write MCQ options grounded to labelled buttons on user image(s), creating a vision-language option space. There were no restrictions on the number of MCQ options, though there should be at least as many MCQ options as there are buttons, with each button referenced in at least one option. The structure of the MCQ options varied by question type. MCQ options for pure functionality questions were typically written as \texttt{action <button>}, such as \textit{Press} \texttt{<button1>}. For specific functionality questions, relevant function details were appended to the option space \textit{i.e.,} \textit{Press} \texttt{<button1>} \textit{to select Defrost}. This evaluated the agent's ability to reason about the user's desired outcome. To push the model to understand the underlying semantics of QA pairs, at least one wrong MCQ option had similar instructions to the correct one.

In summary, 18 participants were engaged in video collection; over 350 man-hours were involved in recruitment, filming and liaison over a span of 6 months. 4 annotators were then engaged over another 300 man-hours to label images and create multiple-choice, multi-step QA pairs.

\subsection{Dataset Statistics}

\begin{figure*}[t]
\centering
\subfigure[]{
\includegraphics[width=0.475\linewidth]{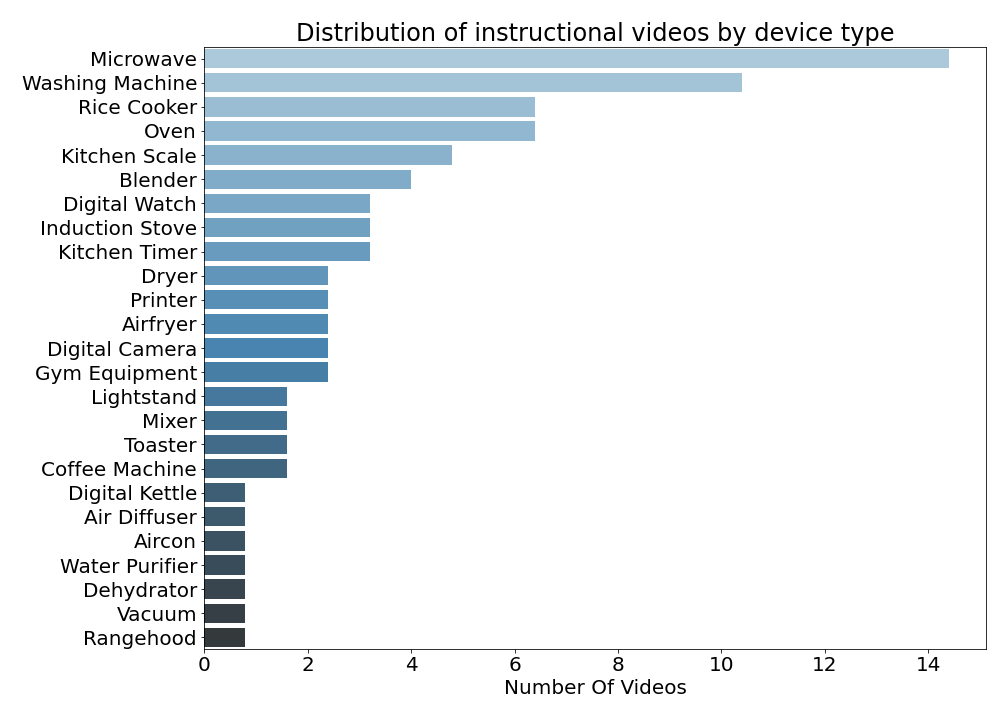}
}
\subfigure[]{
\includegraphics[width=0.475\linewidth]{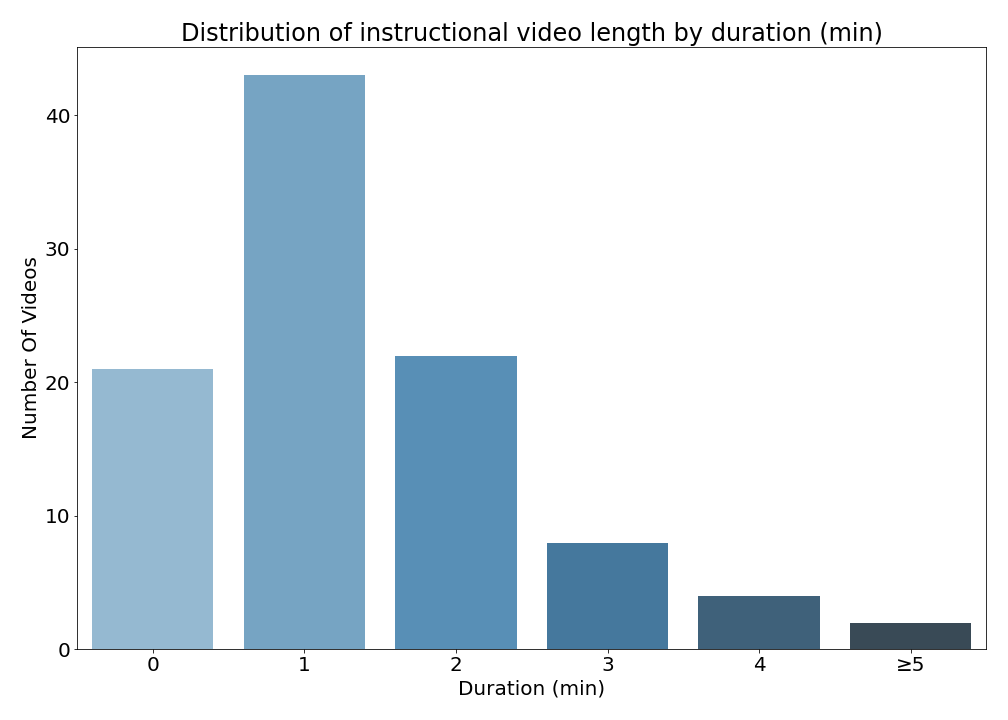}
}
\caption{Statistics of Videos in AssistQ. (a) Distribution of devices by appliance types. (b) Distribution of instructional videos by length.}\label{Video Collection}
\end{figure*}

\begin{figure*}[t]
\centering
\subfigure[]{
\includegraphics[width=0.475\linewidth]{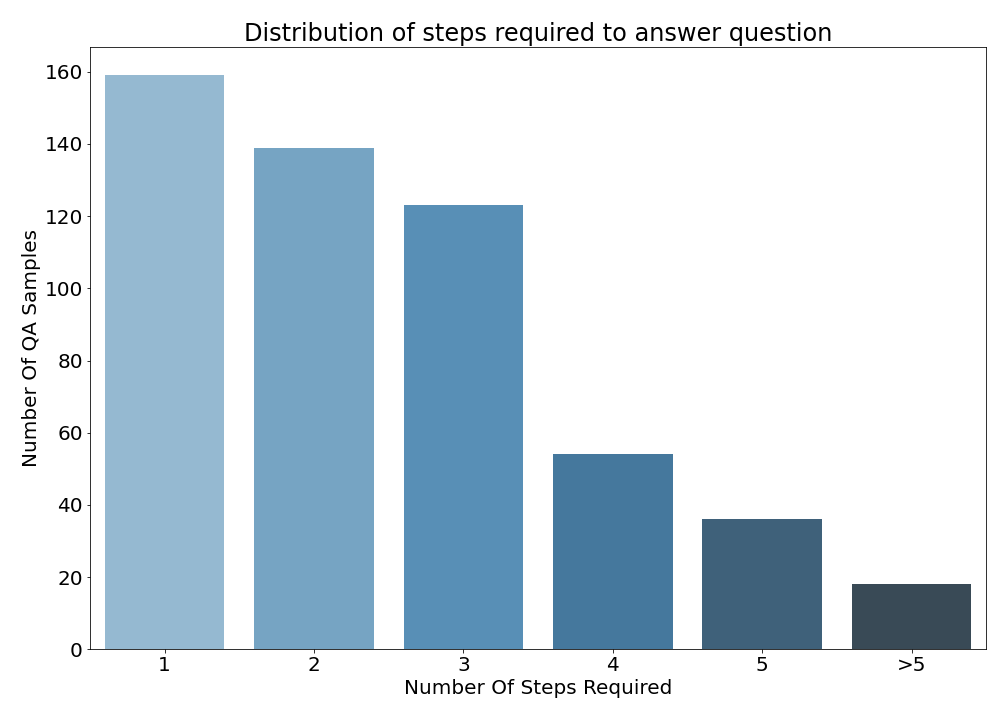}
}
\subfigure[]{
\includegraphics[width=0.475\linewidth,height=0.35\linewidth]{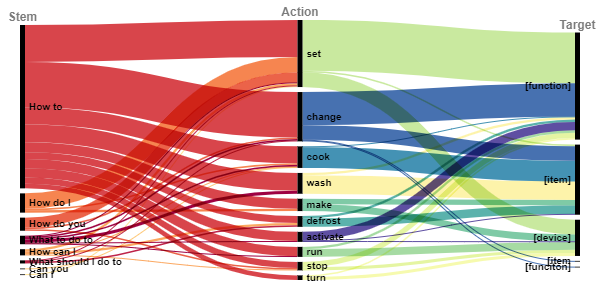}
}
\caption{Statistics of Questions in AssistQ. (a) Distribution of steps required to answer generated queries. (b) Analysis of question structure by stem-action-target.}\label{Question Collection}
\end{figure*}

100 videos with duration averaging 115 seconds were collected. Figure~\ref{Video Collection}(b) shows the distribution of the video duration, with instructional videos spanning 1-2 min having the highest frequency in AssistQ. The videos spanned 25 common household appliances, with microwave, washing machine and oven having the highest frequencies due to their ubiquity in households (Figure~\ref{Video Collection}(a)). The devices came from a diverse pool of 53 distinct appliance brands such as Panasonic (5\%), Cornell (4\%), Sharp (4\%), LG (4\%) and Bosch (4\%). Instructional videos of more complex devices such as printers (240s) and gym equipment (216s) have longer average duration to capture their larger variety of functions.

From the 100 instructional videos, we collected 531 multiple-choice QA pairs (Figure~\ref{Question Collection}) Of which, 251 are pure functionality questions and 278 are specific functionality questions. 70\% of QA pairs require more than 2 steps in the answer sequence (Figure~\ref{Question Collection}(a)), rendering a majority of our questions multi-step. Figure~\ref{Question Collection}(b) shows the alluvial chart of question stem-action-target words obtained from the generated queries. Only the top 10 most common action words were included in the alluvial chart for greater clarity. From the chart, we can see that while distributed, questions beginning with \textit{``How to"} and \textit{``How do I"}, actions words \textit{set} and \textit{change} as well as target words \textit{[function]} and \textit{[item]} having the highest occurrence in our dataset. Apart from diversity, this dataset also poses a large number of interesting challenges. See supplementary material for details.

\section{Question-to-Actions (Q2A) Model}\label{model}

\begin{figure*}[t]
\centering
\includegraphics[width=0.95\linewidth]{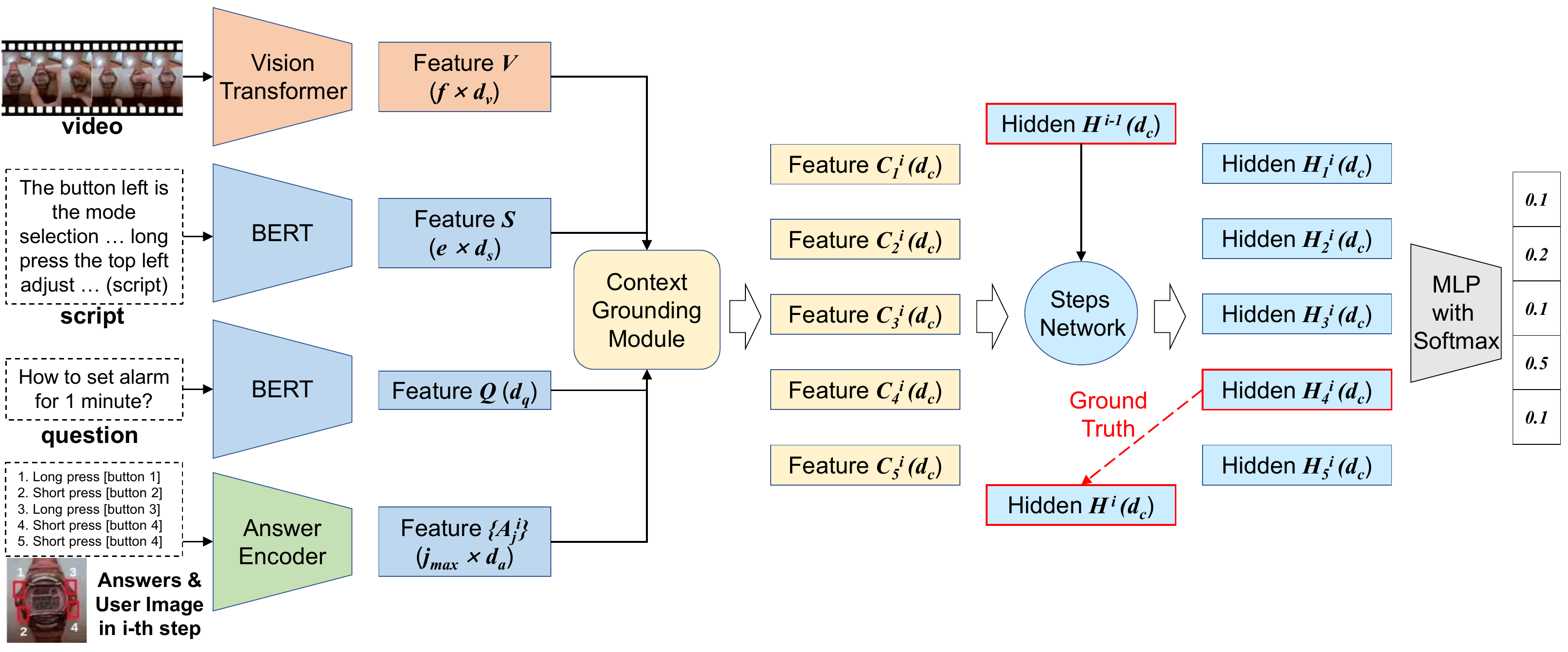}
\caption{Architecture of our question-to-answer (Q2A) model. It can be viewed as an encoder-decoder architecture, where the input encoders obtain feature representations for different inputs (video, script, question, answer), and the decoder modules (context grounding, steps network) estimate the answer from encodings. Best viewed in colors.}\label{figure:model}
\end{figure*}

In this section, we propose a question-to-actions (Q2A) model for the proposed AQTC task. As shown in Figure~\ref{figure:model}, our Q2A model is an encoder-decoder architecture including: (1) Input Encoder extracts feature from video, script, question, and answers; (2) Context Grounding Module generates context-aware feature for each question-answer pair; (3) Steps Network answers the question in the current step. Next, we describe them in detail.

\subsection{Input Encoders}\label{section5.1}

\noindent\textbf{Video.} Video encoding aims to obtain the visual location and of buttons. This not only requires the video embedding~\cite{unstl,videomae,allinone}, but also requires some object information in video embedding~\cite{oatrans,region_learner,higcin}. For simplicity, we just use a pretrained vision transformer (ViT)~\cite{vit} to encode frames instead of a video transformer~\cite{vivit,timesformer}, but use a special button encoding method for user image (described latter). We encode one frame per second for the video. Suppose the video has $f$ frames and the encoding dimension is $d_v$, we can obtain the feature representation $V$ of shape $f \times d_v$.

\noindent\textbf{Script.} The video script describes the button's operation and function, which are the crucial cues to answer the question. We use BERT~\cite{bert} for text embedding. Because the answer in a step usually corresponds to a specific sentence in the script, we transform the script into $e$ sentences, and use the pooled outputs from BERT as sentence embeddings $S$ ($e \times d_s$), where $d_s$ is the feature dimension.

\noindent\textbf{Question.} Following the embedding way of the script, we also use the pooled output from BERT~\cite{bert} as the question feature $Q$\footnote{$Q_U$ in Section~\ref{section3.2} denotes the question under user view. To simplify, we use $Q$ here.}, with the dimension $d_q$. To keep the consistency, we leverage the same BERT used in the script (\textit{i.e.,} $d_q = d_s$). By this way, the question feature and the script feature may have similarities, which can help answer the question. To distinguish the question from the script, we add the prefix ``Question:'' to each question before encoding. 

\begin{figure*}[t]
\centering
\includegraphics[width=0.95\linewidth]{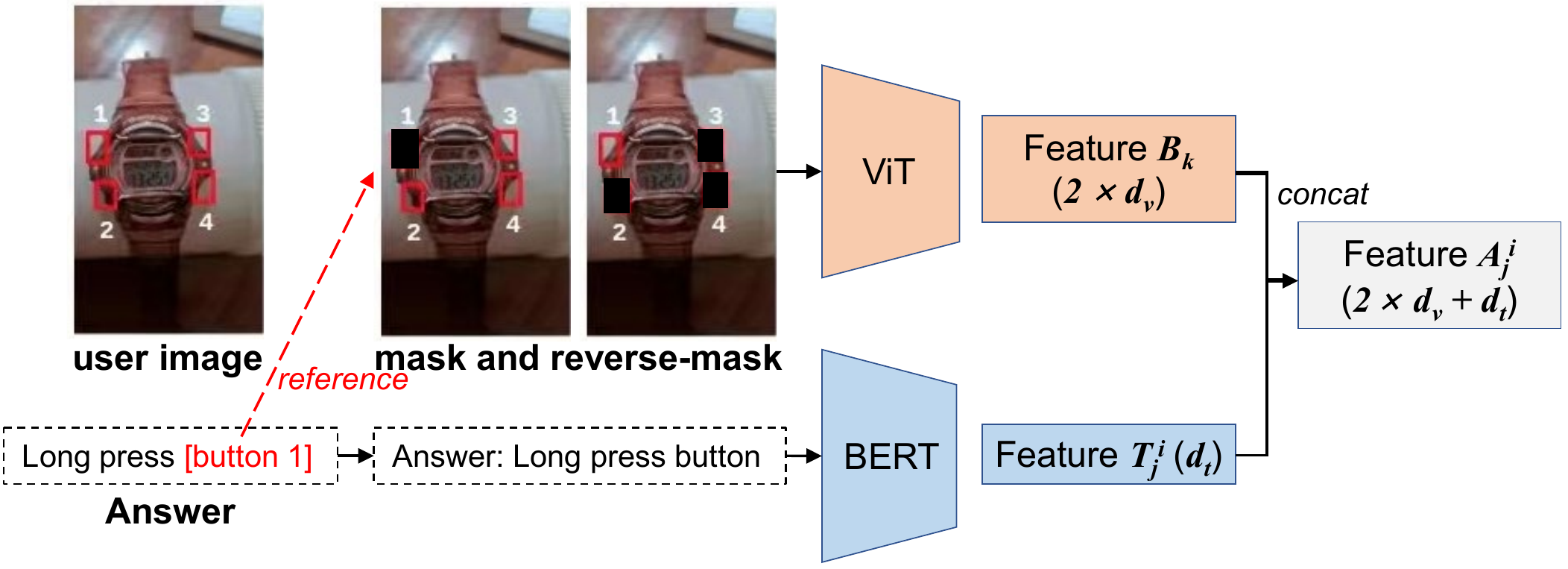}
\caption{The architecture of the answer encoder. This figure presents the case of visual button encoding with both mask and reverse-mask images. Note that ViT and BERT used here are the same networks shown in Figure~\ref{figure:model}. Best viewed in colors.}\label{figure:answer-encoder}
\end{figure*}

\noindent\textbf{The $j$-th answer of the $i$-th step.} An answer contains both text and visual button (bounding-box in the user image), making it difficult to encode. For answer text, we still use BERT to extract the feature. Like the prefix in question, we also prefix each answer with ``Answer:''. The challenging part is the button representation. As shown in Figure~\ref{figure:answer-encoder}, we encode the user image $U$ with the masked referenced button. Meanwhile, we also encode $U$ with all buttons masked except the referenced button. The visual button feature $B_k$, and text feature $T^i_j$ are concatenated to the answer's feature $A^i_j$. The masked button image allows our model to utilize the appearance cues when buttons differ in their appearance, and another reverse-masked image allows our model to focus on the mutual cues (like the relative location) when buttons' appearances are similar.

\noindent\textbf{Context Grounding Module.} In video QA models~\cite{iperceive,tvqa} there are usually modules that utilize context for better question and answer representation. Following this idea, we also design a context grounding module. Consider the input feature representations $V$, $S$, $Q$, ${A^i_j}$ ($T^i_j$ as the text encoding, $B^i_j$ as the button encoding), we generate context-aware question-answer pair feature $C^i_j$ as:

\begin{equation}
    C^i_j = MLP([T^i_j, B^i_j, At_{QA\rightarrow S}(T^i_j, S, S), At_{QA\dashrightarrow S \dashrightarrow V}(T^i_j, S, V)])
\end{equation} 

\noindent  where $At(query,key,value)$ denotes the attention operation~\cite{attention}. $MLP$ denotes a 2-layer MLP, of which dimension is carefully adjusted to satisfy the calculation requirements. $[\cdot]$ is the concatenate operation. $QA \rightarrow S$ denotes the attention from question-answer (query) to script (key), and produces the QA-aware script feature. We also introduce a ``transfer attention'' ($QA \dashrightarrow S \dashrightarrow V$) here to simulate the process of finding visual cues based on text: 

\begin{equation}
    Mask_{QA\dashrightarrow S \dashrightarrow V}(T^i_j, S, V) = Mask_{QA\rightarrow S}(T^i_j, S, S) \times Mask_{S\rightarrow V}(S, V, V),
\end{equation}

\noindent where $Mask$ denotes the attention mask. With the attention mask, we can obtain the QA-aware video feature by $At_{QA\dashrightarrow S \dashrightarrow V} = Mask_{QA\dashrightarrow S \dashrightarrow V} \times V$.

\noindent\textbf{Steps Network.} Our Q2A leverages the historical steps via a steps network, which is a 2-layer MLP or a GRU. In $i$-th step, the steps network would use the state from $i-1$-th step to produce the state $H_j^i$ for $j$-th answer. Then, the state of the ground-truth answer would be reused as the hidden state of the $i+1$ step. The initial state is a random-initialized vector with standard normal distribution if the step just begins. 

\noindent\textbf{Prediction Head.} In each step, the hidden states generated by the steps indicate the answers' feature representations. We use a two-layer MLP followed by softmax activation to predict a score for each answer. 

\noindent\textbf{Training and Inference.} Following common practice in Visual QA/Dialog~\cite{viddial,bottom_up_and_top_down}, BERT and ViT in the input encoder are frozen since they have a large number of parameters. Other parts are trainable. We use cross-entropy (CE) with softmax activation to calculate the loss. During inference, our Q2A model chooses the state that can yield the highest predicted score after the prediction head. In this way, the model can finish the whole multi-step answering procedure by itself.

\section{Experiments}

\subsection{Experimental Setting}

\noindent\textbf{Data Splits.} We randomly split 100 instructional videos of AssistQ into the training set and the validation set with a ratio of 8:2. In statistics, the training set has 80 instructional videos with 425 QA instances, and the testing set has 20 instructional videos with 106 QA instances. About 25\% device types in the testing set are not present in the training set, making our AssistQ benchmark more challenging.

\noindent\textbf{Evaluation Metrics.} As illustrated in Section~\ref{dataset}, the ``multi-step'' characteristic is similar to ``multi-round'' characteristic in visual dialog~\cite{viddial,visdial}. Therefore, it is natural to use their evaluation metrics for the proposed AssistQ task:

$\bullet$ \textbf{Recall@k} measures how often the ground-truth answer selected in top k choices. We report Recall@1 and Recall@3 in our experiments.

$\bullet$ \textbf{Mean rank (MR)} refers to mean value of the predicted ranking position of the correct answer. The model should pursue lower MR.

$\bullet$ \textbf{Mean reciprocal rank (MRR)} is the mean value of the reciprocal predicted ranking position of the correct answer. Higher MRR is better.

\noindent\textbf{Implementation Details.} We use PyTorch~\cite{pytorch} to perform experiments. For a fair comparison, all models use the same optimizer and learning rate scheduler. Specifically, we use a momentum SGD optimizer with a batch size of 16, and a cosine annealing scheduler~\cite{sgdr} with the learning rate $2\times10^{-3}$, 1 epoch for warmup, and maximum 6 training epochs. 

\subsection{Ablation Study}

As illustrated in Section~\ref{model}, we design an encoder-decoder architecture Q2A to solve the AQTC task. In its encoder, we encode the video, script, question, and candidate answers, where a special visual button encoding method is designed for the answer encoding. Meanwhile, in the Q2A decoder, we also specifically develop the context grounding module and utilize historical steps for better prediction. A default setting of our model is (a) encode both video and script, (b) double-mask button encoding, (c) $QA \rightarrow S \rightarrow V$ attention, (d) MLP for historical steps. Next, we perform ablation studies on the points mentioned above.  

\begin{table}[t]
    \centering
    \scriptsize
    \subtable[Ablation on button encoding]{
        \begin{tabular}{c|c|c|c|c|c}
        Mask & Reverse-mask & R@1 $\uparrow$ & R@3 $\uparrow$ & MR $\downarrow$ & MRR $\uparrow$ \\
        \hline
        $\times$ & $\times$ & 17.9 & 54.8 & 3.8 & 2.5 \\
        $\checkmark$ & $\times$ & 19.4 & 53.6 & 3.7 & 2.5 \\
        $\times$ & $\checkmark$ & \textbf{30.2} & 62.3 & 3.2 & 3.2 \\
        $\checkmark$ & $\checkmark$ & 21.4 & 59.1 & 3.7 & 2.6 \\
        \end{tabular}
    }
    \subtable[Ablation on input modalities]{
        \begin{tabular}{c|c|c|c|c|c}
        Video & Script & R@1 $\uparrow$ & R@3 $\uparrow$ & MR $\downarrow$ & MRR $\uparrow$ \\
        \hline
        $\times$ & $\times$ & 23.4 & 57.9 & 3.6 & 2.8 \\
        $\checkmark$ & $\times$ & 29.4 & 59.5 & 3.3 & 3.1 \\
        $\times$ & $\checkmark$ & 24.6 & 58.7 & 3.6 & 2.9 \\
        $\checkmark$ & $\checkmark$ & \textbf{30.2} & 62.3 & 3.2 & 3.2 \\
        \end{tabular}
    }
    \subtable[Ablation on context grounding]{
        \tiny{
        \begin{tabular}{c|c|c|c|c|c|c}
            QA$\rightarrow$S & S$\rightarrow$V & QA$\dashrightarrow$S$\dashrightarrow$V & R@1 $\uparrow$ & R@3 $\uparrow$ & MR $\downarrow$ & MRR $\uparrow$ \\
            \hline
            $\times$ & $\times$ & $\times$ & 19.8 & 52.8 & 3.8 & 2.5 \\
            $\checkmark$ & $\times$ & $\times$ & 28.6 & 69.8 & 3.1 & 3.2 \\
            $\times$ & $\checkmark$ & $\times$ & 22.6 & 56.0 & 3.6 & 2.6 \\
            $\checkmark$ & $\checkmark$ & $\times$ & 29.0 & 62.7 & 3.2 & 3.2  \\
            $\checkmark$ & $\checkmark$ & $\checkmark$ & \textbf{30.2} & 62.3 & 3.2 & 3.2 \\
            \end{tabular}
        }
    }
    \subtable[Ablation on historical steps]{
    \tiny{
        \begin{tabular}{c|c|c|c|c|c}
        Network & History & R@1 $\uparrow$ & R@3 $\uparrow$ & MR $\downarrow$ & MRR $\uparrow$ \\
        \hline
        \multirow{2}{*}{MLP} & $\times$ & 21.4 & 60.7 & 3.5 & 2.7 \\
        & $\checkmark$ & 21.8 & 56.8 & 3.7 & 2.6 \\
        \hline
        \multirow{2}{*}{GRU} & $\times$ & 25.0 & 62.3 & 3.4 & 2.9 \\
        & $\checkmark$  & \textbf{30.2} & 62.3 & 3.2 & 3.2 \\
        \end{tabular}
    }
}
\caption{Ablation studies of our proposed Q2A model on the AssistQ benchmark. (a) and (b) are for the Q2A encoder, where (a) explores the necessary input modalities and (b) studies the ways of button encoding. (c) and (d) are for the Q2A decoder, where (c) illustrates the importance of using historical steps and (d) shows different combinations for context grounding. Note $\rightarrow$ and $\dasharrow$ denote attention methods we described in Section~\ref{section5.1}.}\label{table:encoder-decoder}
\end{table}

\noindent\textbf{Visual Button Encoding.} See Table~\ref{table:encoder-decoder}(a), we analyze the encoding ways of visual button encoding. Interestingly, the model achieves the best performance when only using reverse-mask image encoding (mask $\times$ and reverse-mask $\checkmark$), with a clear margin compared to other configurations. These results demonstrate that it is better to mask the other buttons when encoding a visual button. In our opinion, both mask and reverse-mask schemes can help the model to infer the button location, but the reverse-mask scheme can help the model use the button appearance.

\noindent\textbf{Input Modalities.} As shown in Table~\ref{table:encoder-decoder}(b), when the model ignores some contexts (video $\times$ or script $\times$), it achieves much lower performance (23.4 v.s. 30.2) than the model with the full contexts (video $\checkmark$ and script $\checkmark$). We also found that the independent video (video $\checkmark$ or script $\times$) can lead to much better results (29.4 v.s. 24.6) than the independent script (video $\times$ or script $\checkmark$). This is because when using only script w/o video,
some information in scripts might be misleading. For example,
say we have two candidate answers: (a) “press here
[refer to $<$button1$>$ in video, but not accessible when using
only script] on the microwave”; (b) “press here [refer to
$<$button2$>$] to heat it up”. If the script has a language context
(\textit{e.g.,} “press here to heat it up”) which is similar to (b),
it will mislead the model, making it more inclined to choose
(b); but in fact, (a) is the ground truth option, which we can
clearly tell by watching the instructional video. These results suggest that the model should fully utilize both video and script to answer questions, rather than only one of them.

\noindent\textbf{Context Grounding.}\footnote{We use the same notation of Section~\ref{model}. $V$: video, $S$: script, $Q$: question, $A$: answer.} According to Table~\ref{table:encoder-decoder}(c), all parts in our proposed grounding schemes (see Section~\ref{section5.1}) can improve the performances. Among them, the model benefits less from $S \rightarrow V$ but obtains a significant gain from $QA \rightarrow S$. This is because $S \rightarrow V$ is unrelated to the question, suggesting the importance of finding the corresponding context according to the question. Moreover, based on $QA \rightarrow S$ and $S \rightarrow V$, the model would benefit from the ``transfer attention'' $QA \dashrightarrow S \dashrightarrow V$. We believe the ``transfer attention'' could link $QA$ to $V$, which simulates finding the related script based on the question, and then locating the related video moment by the related script. Compared with direct $QA \rightarrow V$, the calculation cost of our ``transfer attention'' is much smaller since the script is shared for all candidate QA pairs.

\noindent\textbf{Steps Network.} Table~\ref{table:encoder-decoder}(d) presents the ablation on whether to use cues from historical steps ($\checkmark$ or $\times$) and how to use them (MLP or GRU~\cite{gru}). We can find that historical information can stably improve the performances of GRU on all metrics, especially on Recall@1 (25.0 to 30.2). But for MLP with historical information, only a few gains (21.4 to 21.8) can be observed on Recall@1, while other metrics even declined. Compared to MLP, GRU has advantages in temporal modeling, which is more suitable as the steps network.

\begin{table}[t]
\centering
\begin{tabular}{c|c|c|c|c}
Baseline Method & Recall@1 $\uparrow$ & Recall@3 $\uparrow$ & MR $\downarrow$ & MRR $\uparrow$ \\
\hline
Random Guess & 18.3 & 50.4 & 3.9 & 2.3 \\
\hline
Multi-stream~\cite{tvqa} (Video QA) & 22.8 (+4.5) & 54.8 (+4.4) & 3.7 (-0.2) & 3.0 (+0.7) \\
LateFusion~\cite{viddial} (Video Dialog) & 19.9 (+1.6) & 49.3 (-1.1) & 4.1 (+0.2) & 2.6 (+0.3) \\
PACMAN~\cite{eqa} (EQA) & 24.6 (+6.3) & 54.0 (+3.6) & 3.9 (-0.0) & 2.9 (+0.6) \\
\hline
Q2A (Ours) & \textbf{30.2 (+11.9)} & \textbf{62.3 (+11.9)} & \textbf{3.2 (-0.7)} & \textbf{3.2 (+0.9)} \\
\end{tabular}
\caption{Baseline comparison on the AssistQ benchmark. Benefiting from the specific design for the benchmark, Q2A achieves the highest performance on all metrics.}\label{table:comparison}
\end{table}

\subsection{Comparison}

As shown in Table~\ref{table:comparison}, we compare our Q2A model with baselines in other domains on the AssistQ benchmark. Since the answers encoded in other methods are plain text (without placeholders for buttons), we simply introduce our double-mask answer encoding scheme. For all models, we use the same backbone network in the encoder for a fair comparison. We also provide the results of random guess for improvement comparison.

\noindent\textbf{Video QA.} Multi-stream is proposed as a baseline model for the TVQA~\cite{tvqa} task. Like our model, it has modules similar to context grounding. However, It cannot take advantage of cues from historical steps (as video QA is a single-step task), limiting its performance on the AssistQ benchmark.

\noindent\textbf{Video Dialog.} \cite{viddial} proposes LateFusion to solve the multi-round video dialog, which simply concatenates the historical dialog and introduces an LSTM~\cite{lstm} to model them. However, its performance is even worse than Multi-stream. We believe that simple late fusion leads to inferior performance as there is no module similar to context grounding. Also, its LSTM models a very long sequence of the entire script and historical dialogue, which may lose useful information.

\noindent\textbf{EQA.} PACMAN~\cite{eqa} uses an LSTM-based planner to update the agent state, with an MLP-based control planner to predict answers. This procedure is similar to our GRU-based decoder that utilizes historical steps, but PACMAN only considers the visual information for navigation. As shown in Table~\ref{table:comparison}, PACMAN achieves better performance than Multi-stream and LateFusion, but still lags behind our Q2A model.

\section{Conclusion}\label{conclusion}
In this paper, we proposed Affordance-centric Question-driven Task Completion (AQTC) task, which enables AI assistants to guide users by learning from instructional videos. To support the task, we collected the AssistQ benchmark, consisting of 531 multiple-step QA samples derived from 100 newly filmed instructional videos, with efforts underway to continue scaling this dataset. We also present a new baseline called Question-to-Actions (Q2A) for our AQTC task, and experimental results demonstrate the effectiveness of the Q2A model on the AQTC task. We hope that our proposed AQTC and AssistQ can advance the development of AI assistants that see the world through our eyes, assisting humans in daily, real-world scenarios.

\noindent\textbf{Acknowledgements.} This project is supported by the National Research Foundation, Singapore under its NRFF Award NRF-NRFF13-2021-0008, and Mike Zheng Shou's Start-Up Grant from NUS. The computational work for this article was partially performed on resources of the National Supercomputing Centre, Singapore.

\clearpage

\appendix
\noindent{\Large{\textbf{Appendix}}}

\section{Dataset Quality Control}

To ensure the quality of videos and scripts, we worked with participants closely (\textit{e.g.} dissemination of comprehensive submission guidelines, checking of device before filming, hands-on recording guidance, review of transcripts). Researchers also conduct a quality check on the recorded videos before acceptance. In fact, 2-3\% of videos have been rejected due to undesired quality (e.g. unclear video instructions). To ensure the quality of QA pairs and annotation, AssistQ also adopted all 4 quality control measures referenced in TGIF-QA~\cite{tgifqa}: gold standard annotation to help annotators to understand requirements, rejection and reviews (each video has been audited by at least 1 more person). The high quality of our data can be further confirmed by the high human performances at 95.8\% recall@1.

\section{Dataset Collection Source}
We noticed that many virtual world simulators could help to produce a clean dataset, but we hold on collecting data in the real world. The main reason is that ultimately we want to deploy such AI assistant models in real environments such as smart glasses. To this end, at least we need to have testing data in real environment. Regarding training data, we agree it is a promising research direction to leverage a virtual environment. However, due to the significant domain gap between the existing virtual environments (\textit{e.g.,} AI2Thor~\cite{ai2thor}, AI Habitat~\cite{aihabitat}) and the real env, real-world training data at the scale provided by us is needed to adapt model trained in virtual to real. In this paper's scope, we focus on models trained only on the real environment; we consider it as future work to pre-train in the virtual environment first and then fine-tuning it on real-world data. 

\begin{figure*}[t]
    \centering
    \includegraphics[width=\linewidth]{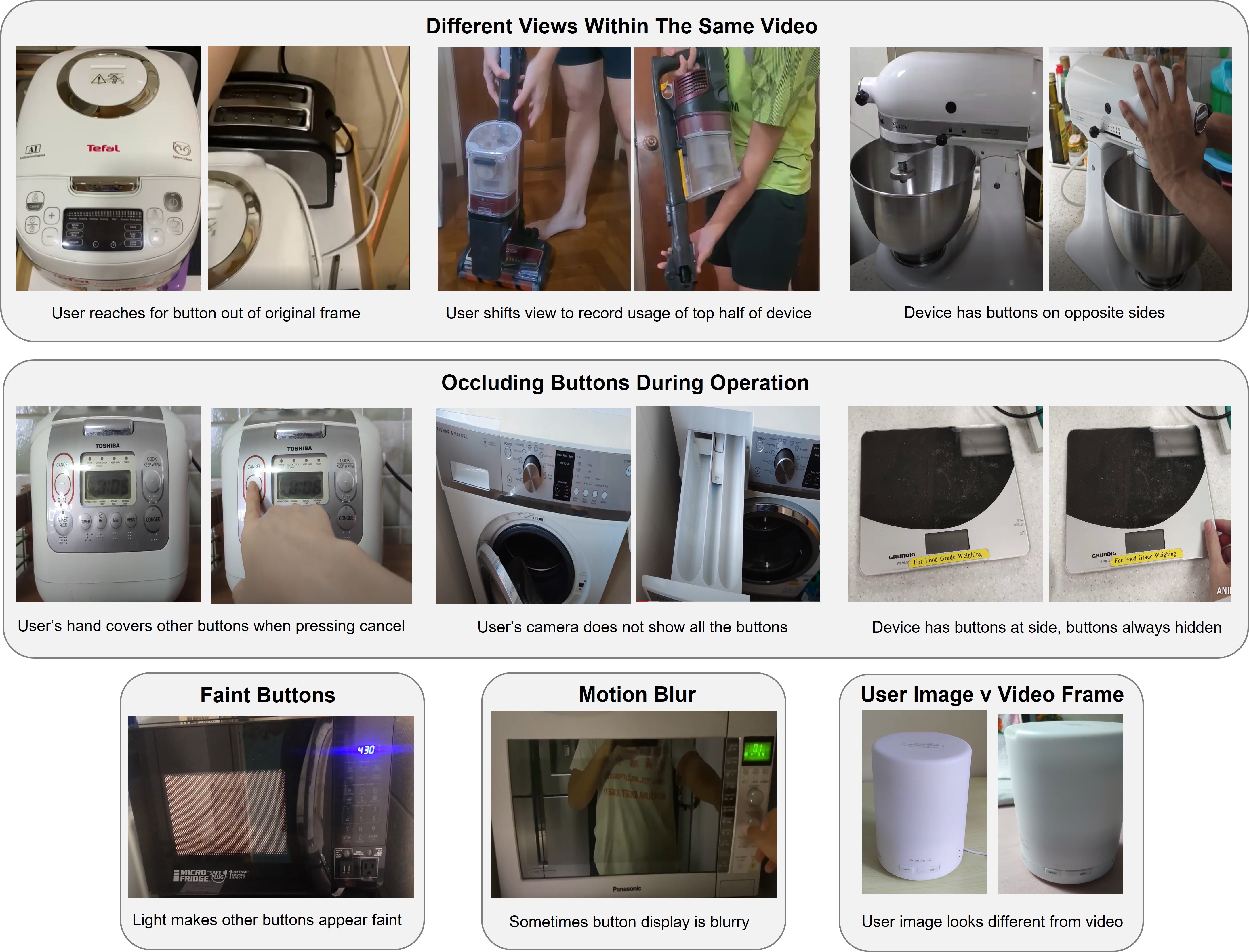}
    \caption{Some challenging examples in our AssistQ datasets.}
    \label{Challenges}
\end{figure*}

\section{Some Dataset Challenges}

There are some interesting challenges in our dataset. These challenges are summarized in Figure~\ref{Challenges}. Some challenges include (a) changing views within the same instructional video; (b) occlusion of buttons due to hand placement, narrow views or hidden buttons; (c) faint buttons from lighting; (d) motion blurs from ego-motion or out-of-focus cameras; as well as; (e) user images taken under different angles and lighting from the video, reiterating the richness and complexity of AssistQ.

\section{Dataset Comparison}

Table~\ref{compare_datasets} compares AssistQ to related datasets. As we can see, AssistQ is unique compared to others: (1) the video content is sourced from real-world, everyday situations in the egocentric perspective, (2) the question are affordance-centric \textit{i.e.} how to use and execute functions of appliances, and multi-modal to more closely reflect the inputs that an AI assistant (\textit{e.g.} AR glass) receives from the user. (3) Answers are a sequence of actions for the user to execute. AssistQ also has longer videos on average compared to most datasets.

\begin{table}[h]
\resizebox{\textwidth}{!}{%
\begin{tabular}{@{}l|l|l|l|c|c|l|c|c|c@{}}
\multirow{2}{*}{Dataset} & \multirow{2}{*}{Video Source} & \multirow{2}{*}{Video Type} & \multirow{2}{*}{Questions} & \multicolumn{2}{c|}{Question Modality} & \multirow{2}{*}{Answers} & \multirow{2}{*}{\#Clips} & \multirow{2}{*}{\#QA} & \multirow{2}{*}{Ave Dur (s)} \\ 
 & & & & Text & Visual & & & & \\ 
\hline
MovieQA~\cite{movieqa} & Movie & Third-person & Factoid & \checkmark & - & Single-step  & 408 & 14,944 & 202.7\\
TVQA~\cite{tvqa} & TV show & Third-person & Factoid & \checkmark & - & Single-step & 21,793 & 152,545 & 76.2\\
AVSD~\cite{viddial} & Crowdsourced & Third-person & Factoid & \checkmark & - & Multi-round & 11,816 & 118,160 & 30 \\
Embodied QA~\cite{eqa} & Simulated env. & Egocentric & Factoid & \checkmark & - & Multi-step & - & 5,281 & -  \\
\textbf{AssistQ (ours)}  & Crowdsourced & Egocentric & Affordance & \checkmark & \checkmark & Multi-step & 100 & 531 & 115\\ 
\end{tabular}}
\caption{Comparison of AssistQ with related datasets.} \label{compare_datasets}
\end{table}

\begin{figure*}[h]
\centering
\includegraphics[width=\linewidth]{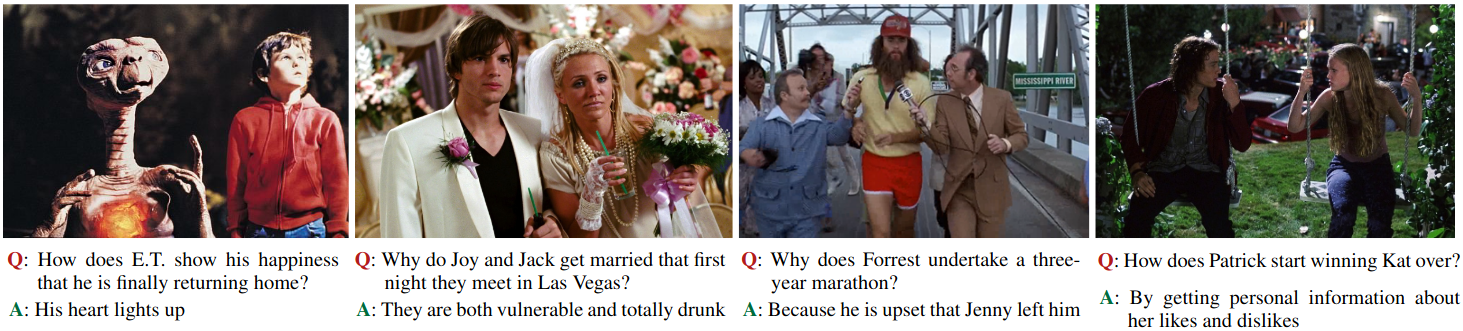}
\caption{Examples from the MovieQA dataset. 18\% of questions were about \textit{who}, followed by 12.4\% about \textit{why} and 9.7\% about \textit{what}. As seen in the examples, questions typically focused on plot developments with little reference to visual signals.}
\label{figure:movieqa}
\end{figure*}

\textbf{MovieQA~\cite{movieqa}.} The goal of MovieQA is to understand story plots in movies. 408 subtitled movies were collected together with their \textit{Wikipedia} synopsis and \textit{imsdb}/Described Video Service (DVS) scripts where available. In the first round of annotation, the annotators were shown the plot synopsis only and asked to create any number of QA pairs that can be localized to a set of sentences in the synopsis. Naturally, this led to questions that were plot-focused and less reliant on visual information. In the second round of annotation, annotators were asked to create 5 multiple-choice answers (1 right, 4 wrong) based on the synopsis and questions. Finally, each sentence of the synopsis was aligned to time-stamps on the video clips ($\sim200$s in length); the video clip and aligned QA pairs then formed the benchmark (Figure~\ref{figure:movieqa}).

\begin{figure*}[h]
\centering
\includegraphics[width=\linewidth]{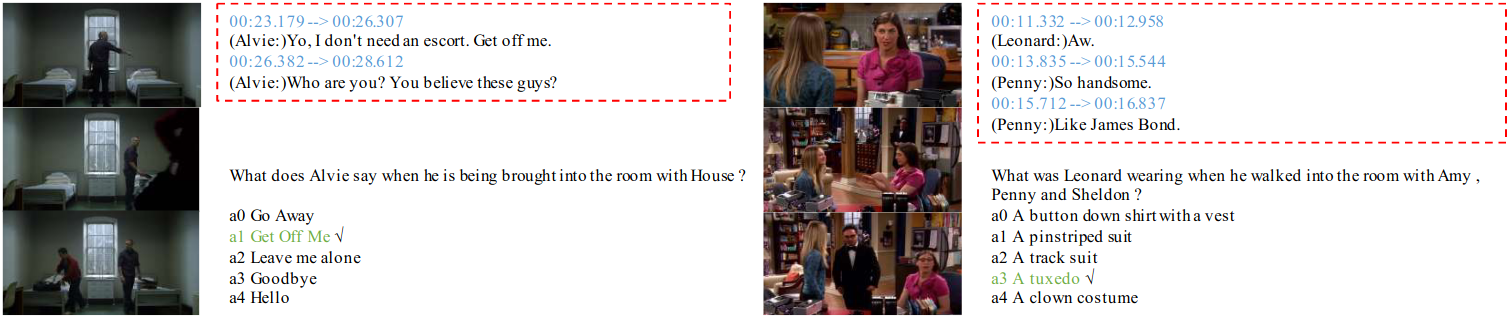}
\caption{Examples from the TVQA dataset. The question had to be written with a [\textit{when/before/after}] clause so that it is localized to a specific moment in the clip. 54\% of questions were about \textit{what}, followed by 15\% about \textit{who}.}
\label{figure:tvqa}
\end{figure*}

\textbf{TVQA~\cite{tvqa}.} Similar to MovieQA, TVQA was created to understand human-centric plots in videos. 6 TV shows were segmented into 60/90-second clips, accompanied by subtitles and aligned transcripts. Annotators were shown the video clip and aligned subtitles, and encouraged to create questions in a 2-part format: [What/How/Where/Why/\textellipsis] \underline{\quad\quad\quad\quad\quad} [when/before/after] \underline{\quad\quad\quad\quad\quad}. The second part served to localize the question to the relevant moment in the clip and ground the question in visual signals, such as \textit{What was House saying before he leaned over the bed?}. Annotators provided 5 multiple-choice answers (1 right, 4 wrong) and annotated time-stamps of the exact video portion required to answer the question (Figure~\ref{figure:tvqa}).

\begin{figure*}[h]
\centering
\includegraphics[width=\linewidth]{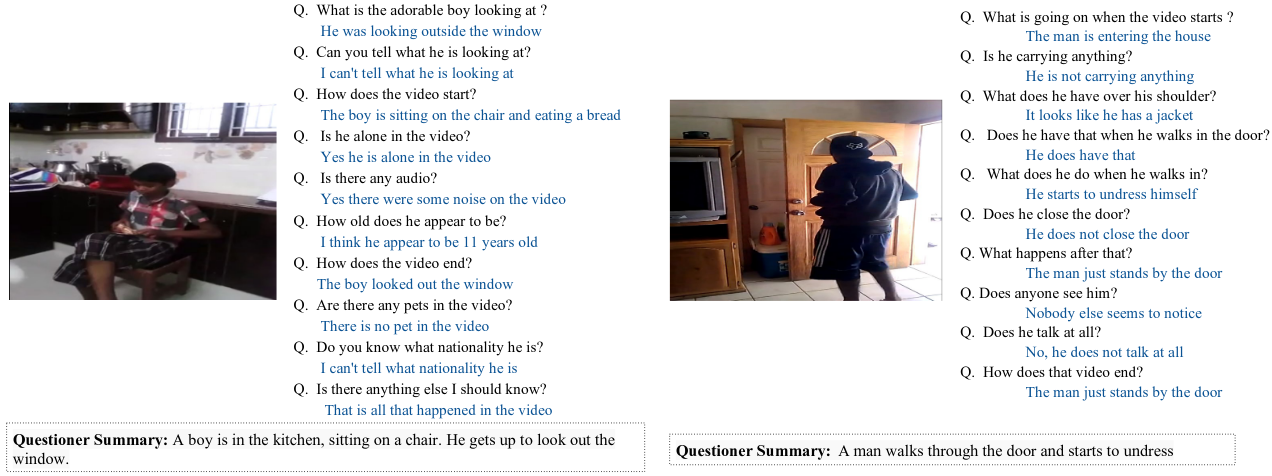}
\caption{Examples from the AVSD dataset. The ``Questioner'' asks a series of questions about the input video and the ``Answerer'' gives an answer at each round. }
\label{figure:avsd}
\end{figure*}

\textbf{AVSD~\cite{viddial}.} In Audio Visual Scene-Aware Dialog (AVSD), an agent is given an input video, a dialog history and a follow-up question, and its goal is to generate a correct response (Figure~\ref{figure:avsd}). 11,816 videos of everyday human activities were taken from the Charades human-activity dataset~\cite{charades}. Each video was handed to a pair of annotators. The ``Questioner'' was tasked to ask questions about activities and events in the video clip, given only 3 video frames. The  ``Answerer'' has access to the video and script, and answers the ``Questioner'' over a sequence of 10 questions. Once the conversation is complete, the ``Questioner'' is tasked to summarize the video.

\begin{figure*}[h]
\centering
\includegraphics[width=\linewidth]{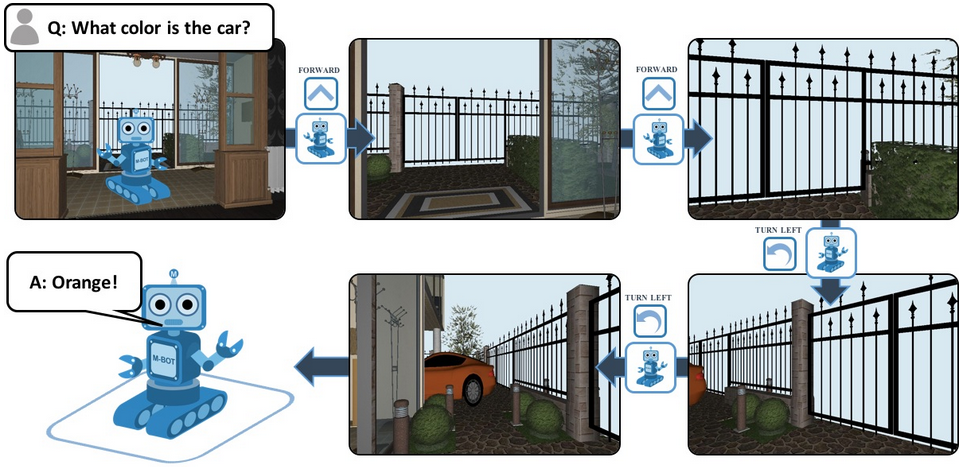}
\caption{Example from the EQA dataset. The agent is spawned at a random location in a simulated home environment and navigates the environment to answer a question about the location/colour of an unambiguous object.}\label{figure:eqa}
\end{figure*}

\textbf{Embodied QA~\cite{eqa}.} In Embodied QA, an agent is spawned at a random location in a simulated home environment and asked a question about the colour/location of an object. The objective of the agent is to navigate the environment through atomic actions (move forward, turn, \textit{etc.}) and gather visual information to answer the question (Figure~\ref{figure:eqa}). The dataset is built on a subset of House3D environments, and the questions were written in specific formats to ensure they were answerable and unambiguous. For example, \textit{location} questions were written as \textit{What room is the \texttt{<OBJ>} located in?}, where \texttt{<OBJ>} is an unambiguous object that is query-able.

\section{Data Sample Visualisation}
We attach an annotated example and an animated video to showcase the the intended use-case of AssistQ in AI assistants. The video shows 2 examples with egocentric instructional videos, and an example with a third-person instructional video.

\begin{figure*}[t]
\centering
\includegraphics[width=\linewidth]{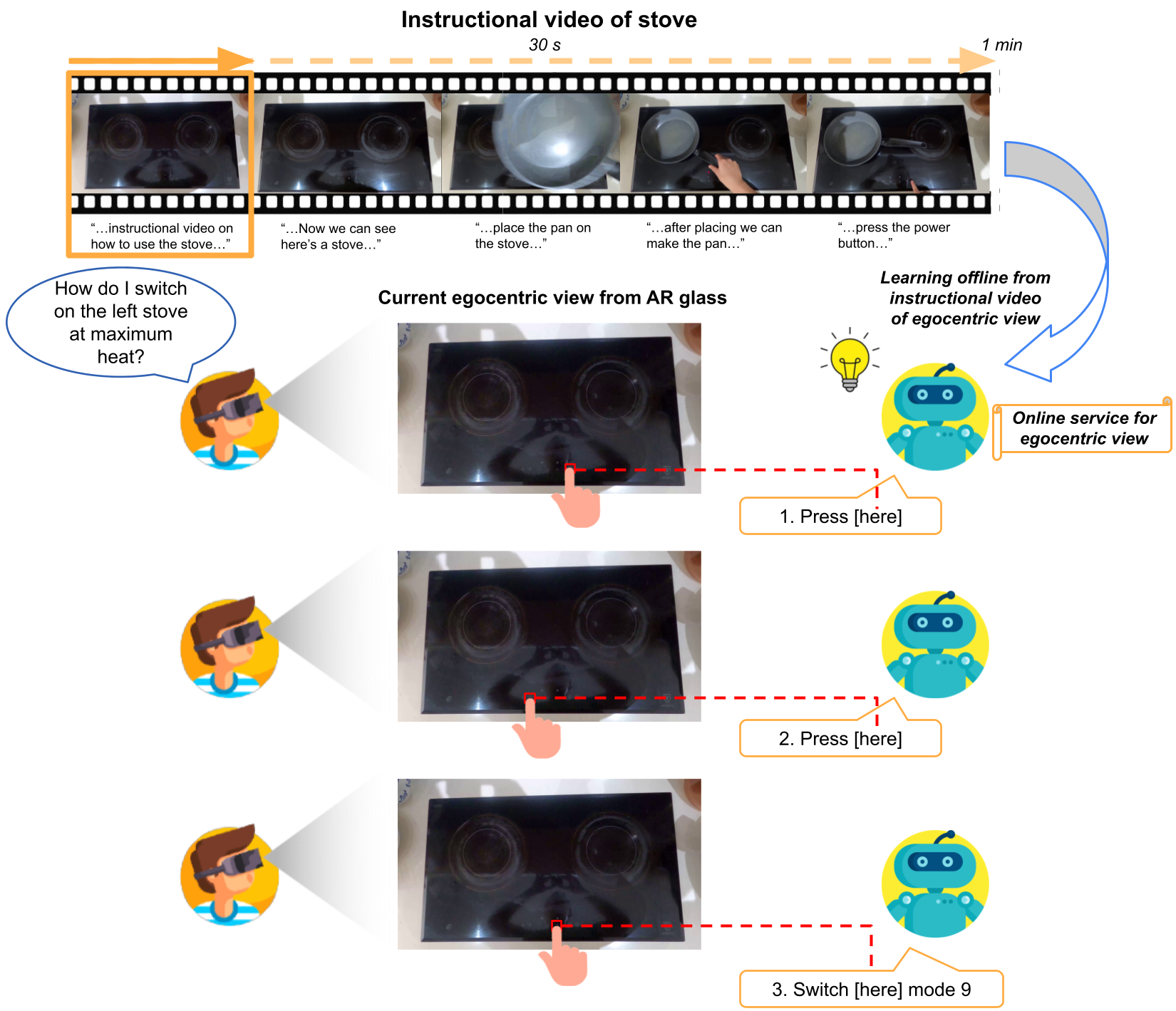}
\caption{Illustration of AQTC task with an example from AssistQ. The AI assistant learns offline from a narrated instructional video. The user wonders how to use the appliance and shows the AI assistant their view. The AI assistant predicts the sequence of actions that the user should perform to accomplish the task, and guides the user through each step with bounding boxes over their view of the appliance.}\label{figure:ego_animated}
\end{figure*}

\textbf{Annotated example.} We provide 1 annotated example from AssistQ to illustrate the format and information encapsulated in our annotations. The example contains the instructional video (video.mp4), question-answering pairs (qa.json), bounding box coordinates (buttons.csv), narration transcript (script.txt) as well as the image folder containing front-view image(s) of the device.

\textbf{Examples with egocentric videos.} We showcase 2 examples from our AssistQ dataset, namely the \href{https://youtu.be/SWIll_d9QH4?t=12s}{EF stove} and \href{https://youtu.be/SWIll_d9QH4?t=1m46s}{Bosch washing machine}, in an animated video (Figure~\ref{figure:ego_animated}). Through these examples, we demonstrate the use of AssistQ in our proposed Affordance-centric Question-driven Task Completion (AQTC) task. The AI assistant learns offline from an egocentric instructional video so that when the user asks a question (e.g. \textit{stove:} How do I switch on the left stove at maximum heat), the AI assistant provides the user with a series of steps to perform (e.g. \textit{stove:} 1. Press \texttt{[here]}, 2. Press \texttt{[here]} and 3. Select \texttt{[here]} mode 9). Each step is grounded to bounding boxes in the user's image, \textit{i.e.} \texttt{[here]} refers to a labelled button/knob on the device. The bounding box is shown on the user's view of the device (\textit{e.g.} through AR glasses) so the user knows the exact action to perform.

\begin{figure*}[t]
\centering
\includegraphics[width=\linewidth]{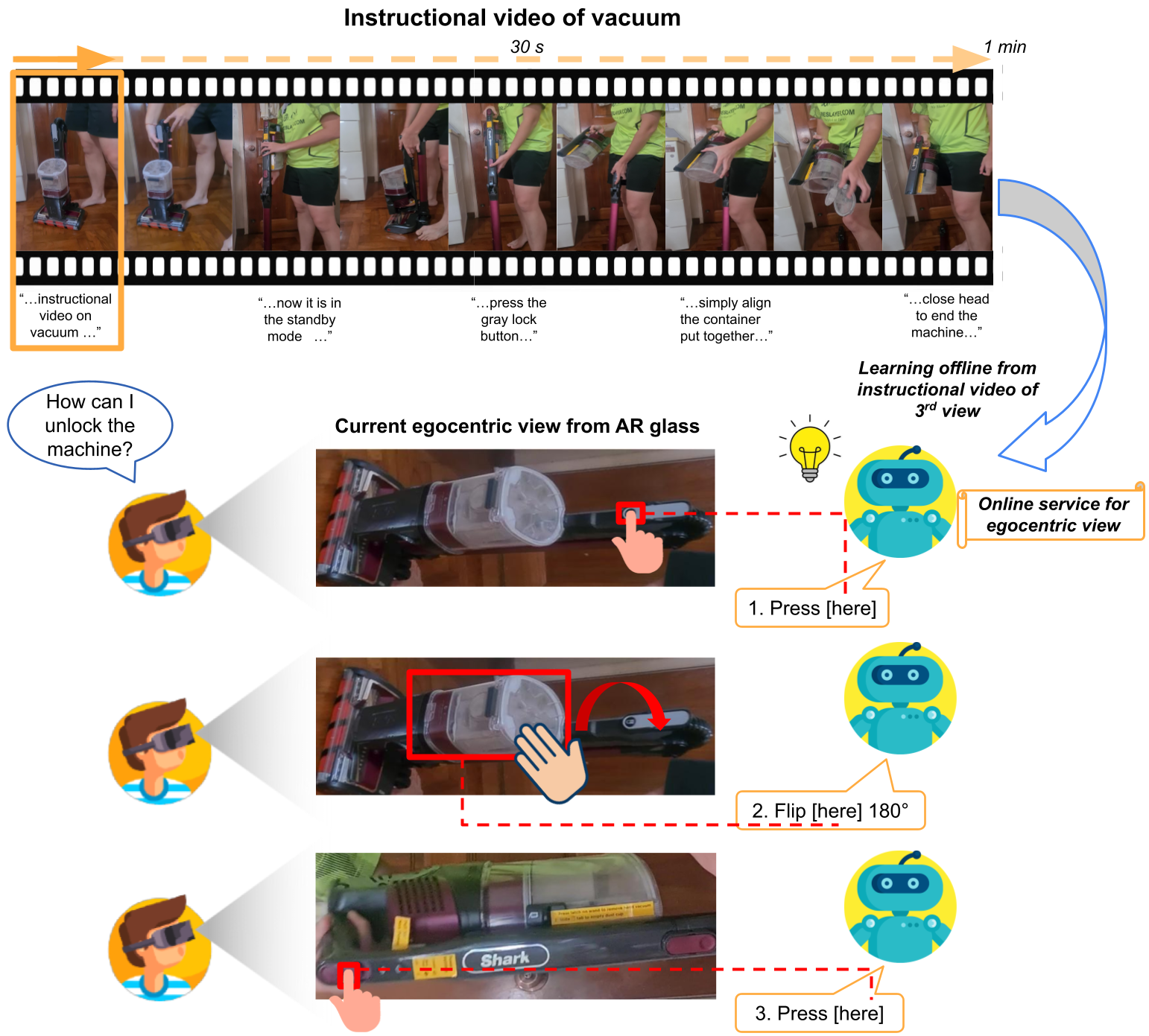}
\caption{Illustration of AQTC task with instructional video recorded from the third-person perspective. The task remains largely unchanged, except that the model has to resolve visual signals from the third-person to the first-person image from the user. This presents new challenges for the AQTC task and we are expanding the AssistQ dataset to include more of such scenarios.}\label{figure:3rdperson_animated}
\end{figure*}

\textbf{Example with third-person video.} In the animated video, we also include a \href{https://youtu.be/SWIll_d9QH4?t=59s}{Shark vacuum} example that had its instructional video recorded in third-person (Figure~\ref{figure:3rdperson_animated}). Compared to the egocentric examples, the third-person example is more challenging as it requires the AI assistant to resolve visual information from the third-person video with that of the user's egocentric visual query. The example demonstrates that the AQTC task need not be confined to egocentric videos. We intend to extend the AssistQ dataset to include more examples with third-person videos, as it is more common for product videos/demos to be recorded in third-person. This will allow further research and development in AI assistants to benefit from our AQTC task and AssistQ dataset.

\bibliographystyle{splncs04}
\bibliography{egbib}

\end{document}